# Reinforcement Learning for Robotic Time-optimal Path Tracking Using Prior Knowledge

*Jiadong Xiao，Lin Li，Yanbiao Zou and Tie Zhang*

School of Mechanical and Automative Engineering, South China University of Technology, Guangzhou, China

**Abstract:**
Time-optimal path tracking, as a significant tool for industrial robots, has attracted the attention of numerous researchers. In most time-optimal path tracking problems, the actuator torque constraints are assumed to be conservative, which ignores the motor characteristic; i.e., the actuator torque constraints are velocity-dependent, and the relationship between torque and velocity is piecewise linear. However, considering that the motor characteristics increase the solving difficulty, in this study, an improved Q-learning algorithm for robotic time-optimal path tracking using prior knowledge is proposed. After considering the limitations of the Q-learning algorithm, an improved action-value function is proposed to improve the convergence rate. The proposed algorithms use the idea of reward and penalty, rewarding the actions that satisfy constraint conditions and penalizing the actions that break constraint conditions, to finally obtain a time-optimal trajectory that satisfies the constraint conditions. The effectiveness of the algorithms is verified by experiments.

## 1. Introduction

The research on the time-optimal path tracking for robots began in 1970 s[1], which is a significant field of industrial robots. The research aims to maximize the performance of the servo motor, to make the robot work at the maximum velocity under the constraint conditions, reduce the execution time for the robotic tasks and improve the working efficiency of the robot, which has important research significance.

The task path of the time-optimal path tracking problem is preset, and the only optimization objective is to optimize the scalar function $t \rightarrow s(t)$, which represents the "position" on the path at each time instant[2]. According to the optimize approaches, the time-optimal path tracking methods can be divided into three groups:

1) Numerical integration [2-7]: The first group of the methods obtain the solution by numerical integration in a way which maximizes the path velocity. The method that using numerical integration to obtain time-optimal trajectory was first proposed in [3]. In [4], the manipulator dynamics were described using parametric functions which represent geometric path constraints to be honored for collision avoidance as well as task requirements and constraints on the input torques/forces are converted to those on the parameters. In [7], the dynamic singularities which are occurred when the velocity limit curve is nondifferentiable is mentioned and a modified algorithm of obtaining the time-optimal trajectory along specified paths considering singularity points is proposed. In [5], a new method which significantly improves the computational efficiency of time-optimal path tracking planning algorithms with limited actuator torques is proposed. In [6], the numerical integration time-optimal path tracking method is used for CNC machining. In [2], a complete solution to the issue of singularities is given.

2) Convex optimization[8-12]: The second group of the methods uses convex optimization techniques to solve the minimum time optimization problems. In [8], a log-barrier-based solution method and a recursive formulation is used to enable online optimization, while in [9] the problem is formulated as a second-order cone program. In [10], based on the work of [9], the nonlinear effects such as viscous friction are considering, which lead to a nonconvex optimal control problem and is solved by a sequential convex programming methods. In [11], the concepts of virtual change rate of the torque and the virtual voltage are introduced and the computationally challenging non-convex minimum time path tracking problem is reduced to a convex optimization problem which can be solved efficiently. In [12], the sequential convex log barrier method is proposed, unlike sequential convex programming, the sequential convex log barrier method linearizes only the concave part of the inequality constraints and subsequently appends all inequality constraints as a weighted logarithmic barrier to the objective.

3) Dynamic programming[13-16]: The third group of methods uses dynamic programming following the idea of Bellman[17]. The idea of using dynamic method to solve the time-optimal problem was first proposed in [13], where the dynamic programming method is used to find the positions, velocities, accelerations, and torques that minimize cost. In [14], three performance criteria: time, the square of velocity, and joint torques are considered and combined by weighting coefficients, and the additional optimizing criteria are studied by applying Bellman's principle. In [16], the torque rate limit and jerk limit are included in the dynamic programming framework which extends the state space to three variables. In [15], a new dynamic programming algorithm was proposed, which uses a suitable interpolation in the phase plane to generate trajectories with continuous joint accelerations and torques.

In most of the earlier mentioned researches, the actuator torque constraints are assumed to be conservative, which has facilitated the obtaining of maximum velocity curve (MVC, see [2]). However, if we want to fully utilize the actuator performance, the servo motor torque characteristic which is velocity-dependent should be considered. As the relationship between servo motor torque and velocity is a piecewise linear function, it is hard to obtain the MVC under considering motor torque characteristic. Therefore, a method should be considered which considering the motor torque characteristic and doesn't need to obtain the MVC.

Recently, some scholars have proposed a reinforcement learning method for autonomous vehicles time optimal velocity control [18], which bring us inspiration that whether it is feasible for robotic time-optimal path tracking by using reinforcement learning. Reinforcement learning is a kind of algorithms that use the idea of interacting with the environment and learn from the interactions to find the optimal policy that maximizes a numerical reward signal[19]. Therefore, it is possible to obtain the time-optimal trajectory by using reinforcement learning methods which does not require the solving of specific MVC.

Reinforcement learning was originally used in game theory, control theory, information theory and operations research, etc. With the development of the research, it has been widely used in the field of robot control. In [20], the reinforcement learning scheme incorporated into the free gait generation makes the robot choose more stable states and develop a continuous walking pattern with a larger average stability margin. In [21], a supervised reinforcement learning approach combined with Gaussian distributed state activation was used for autonomous humanoid robot docking. In [22], an improved Q-learning method was used in autonomous mobile robot path planning.

With the development of reinforcement learning, a variety of approaches have been researched. There are mainly two kinds of reinforcement learning algorithms: value-function-based and policy-search methods[23]. Value-function-based methods attempt to find a policy that maximizes the return by maintaining a set of estimates of expected returns for some policy. Q-learning, as a type of value-function-based reinforcement learning, is a famous learning technique related to the principle of reward and penalties, and also the interaction of the robot with the environment. Although Q-learning has shown a successful implementation in many fields, it has limitation too. When the size of the learning environment increases, not only a longer computational time to update the matrices of Q-value and larger adaptive memory matrices are required, but there is also a possibility of not involving the right probabilities in the converged memory matrices. Besides, during the initial stage of exploration, the motion of the agent is completely random, resulting in wastage of computational effort, slower convergence rate and time-consuming[22].

As mentioned above, Q-learning suffers from slow convergence due to the calculation of all possible action states, but convergence can be improved through appropriate initialization of Q-values using different approaches. In this paper, the use of prior knowledge is considered, which has been validated as effective [18, 24]. In addition, the redesign of the action-value function to be more suitable for solving the time-optimal path tracking problem is also considered.

The rest of this paper is organized as follows. In Section 2, the optimization objective and constraint conditions of time-optimal path tracking problem are described, while in Section 3, an overview of the Q-learning algorithm and its limitation are presented. In Section 4, some approaches for setting the reinforcement learning states as well as improving convergence rate are proposed. In Section 5, an improved Q-learning algorithm for robotic time-optimal path tracking using prior knowledge is proposed, besides, after considering the limitations of Q-learning algorithm, an improved action-value function reinforcement learning algorithm is also proposed. In Section 6, the proposed algorithms are implemented in a 6 degrees-of-freedom (6R) industrial robot with the analysis of the result. Lastly, conclusions are drawn in Section 7.

## 2. Constraint conditions and optimization objective

This Section mainly analyses the dynamic model of a robot manipulator and transforms the dynamic model from joint space into parameter space. The kinematic and dynamic constraints are sequentially analysed, and these constraints are also transformed from joint space into parameter space. Finally, the optimization problem is constructed by the optimization objective function and constraint conditions.

2.1. Dynamic model in parameter space

For a manipulator with n degrees of freedom (DOF), the Cartesian configuration space torque equation can be expressed as follows [25]:

$$\boldsymbol{\tau} = \mathbf{M}(\mathbf{q})\ddot{\mathbf{q}} + \mathbf{B}(\mathbf{q})[\dot{\mathbf{q}}\dot{\mathbf{q}}] + \mathbf{C}(\mathbf{q})[\dot{\mathbf{q}}^2] + \mathbf{F}_v\dot{\mathbf{q}} + \mathbf{F}_c\text{sign}(\dot{\mathbf{q}}) + \mathbf{G}(\mathbf{q}) \tag{1}$$

where $\boldsymbol{\tau} \in \mathbb{R}^n$ is the joint torque of the robot, $\mathbf{M} \in \mathbb{R}^{n \times n}$ is a $n \times n$ positive definite mass matrix, $\mathbf{B} \in \mathbb{R}^{n \times n(n-1)/2}$ is a matrix of dimension of $n \times n(n-1)/2$ of Coriolis coefficients, $[\dot{\mathbf{q}}\dot{\mathbf{q}}]$ is an $n(n-1)/2 \times 1$ vector of joint velocity products given by $[\dot{\mathbf{q}}\dot{\mathbf{q}}] = [\dot{\mathbf{q}}_1\dot{\mathbf{q}}_2 \ \dot{\mathbf{q}}_1\dot{\mathbf{q}}_3 \cdots \dot{\mathbf{q}}_{n-1}\dot{\mathbf{q}}_n]^T$, $C \in \mathbb{R}^{n \times n}$ is an $n \times n$ matrix of centrifugal coeffcients, and $[\dot{\mathbf{q}}^2]$ is an $n \times 1$ vector given by $[\dot{\mathbf{q}}_1^2 \ \dot{\mathbf{q}}_2^2 \cdots \dot{\mathbf{q}}_n^2]^T$, $\mathbf{F}_v \in \mathbb{R}^n$ is a vector of viscous friction parameter, $\mathbf{F}_c \in \mathbb{R}^n$ is a vector of viscous friction parameter, $\mathbf{G}(\mathbf{q}) \in \mathbb{R}^n$ is the gravitational force vector, $\mathbf{q} \in \mathbb{R}^n$ is a vector of the joint angle, $\dot{\mathbf{q}}$, $\ddot{\mathbf{q}}$ is used to denote the first and second derivative of the joint angles with respect to time.

A scalar function $s(t)$ is used to express a robot's pseudo-displacement at time $t$. Therefore, we consider a path $\mathbf{q}(s)$ given in joint space coordinates as a function of $s$, whereas the trajectory's time dependency follows from the relation $s(t)$ between the path pseudo-displacement $s$ and time $t$. Without loss of generality, it is assumed that the trajectory starts at $t = 0$ and ends at $t = T$, so we have $s(0) = 0 \leq s(t) \leq 1 = s(T)$. In addition, since $s \in [0,1]$ and the optimization objective is to minimize the execution time, that is, to make the trajectory as fast as possible, we consider $\dot{s}(t) \geq 0$ almost everywhere for $t \in [0,T]$.

For planning convenience, the joint velocity $\dot{\mathbf{q}}(s) = d\mathbf{q}(s)/dt$ and acceleration $\ddot{\mathbf{q}}(s) = d^2\mathbf{q}(s)/dt^2$ can be rewritten using the chain rule as:

$$\dot{\mathbf{q}}(s) = \mathbf{q}'(s)\dot{s} \tag{2}$$

$$\ddot{\mathbf{q}}(s) = \mathbf{q}'(s)\ddot{s} + \mathbf{q}''(s)\dot{s}^2 \tag{3}$$

where $\dot{s} = ds/dt$ is named the pseudo-velocity, $\ddot{s} = d^2s/dt^2$ is named the pseudo-acceleration, $\mathbf{q}'(s) = \partial \mathbf{q}(s)/\partial s$ is named the pseudo-curvature, which can be used to indicate the smoothness of the path, and $\mathbf{q}''(s) = \partial^2 \mathbf{q}(s)/\partial s^2$ is named the change rate of the pseudo-curvature. The dynamics in joint space are transformed into the dynamics in parameter space by substituting equations (2) and (3) into equation (1), resulting in the following expression:

$$\boldsymbol{\tau}(s) = \mathbf{m}(s)\ddot{s} + \mathbf{c}(s)\dot{s}^2 + \mathbf{f}(s)\dot{s} + \mathbf{g}(s) \tag{4}$$

where

$$\mathbf{m}(s) = \mathbf{M}(\mathbf{q}(s))\mathbf{q}'(s) \tag{5}$$

$$\mathbf{c}(s) = \mathbf{M}(\mathbf{q}(s))\mathbf{q}''(s) + \mathbf{B}(\mathbf{q}(s), \mathbf{q}'(s))\mathbf{q}'(s) + \mathbf{C}(\mathbf{q}(s))(\mathbf{q}'(s))^2 \tag{6}$$

$$\mathbf{f}(s) = \mathbf{F}_v(\mathbf{q}(s))\mathbf{q}'(s) \tag{7}$$

$$\mathbf{g}(s) = \mathbf{F}_c(\mathbf{q}(s))\text{sgn}(\mathbf{q}'(s)) + \mathbf{G}(\mathbf{q}(s)) \tag{8}$$

where $\text{sgn}(\dot{\mathbf{q}}(s))$ is replaced by $\text{sgn}(\mathbf{q}'(s))$ using equation (2) and the assumption that $\dot{s} \geq 0$ almost everywhere in the phase plane $s - \dot{s}$.

2.2. Constraint conditions

*Torque constraint*

For the purpose of operating the robot manipulator at the maximum allowed speed without damaging the components of the robot manipulator, the robot manipulator should be operated at a safe torque. According to equation (4) and the motor torque characteristic, the torque constraint inequality equation can be expressed as:

$$\tau_{\min}(s,\dot{s}) \leq \mathbf{m}(s)\ddot{s} + \mathbf{c}(s)\dot{s}^2 + \mathbf{f}(s)\dot{s} + \mathbf{g}(s) \leq \tau_{\max}(s,\dot{s}) \tag{9}$$

where $\tau_{\min}(s,\dot{s})$ and $\tau_{\max}(s,\dot{s})$ are the torque constraints obtained from the motor torque characteristic.

*Velocity constraint*

To increase the life of the servo motor, the servo motor should be operated under the rated speed as much as possible, and the motor speed should not exceed the maximum allowed speed. According to the maximum allowed speed, we have the maximum motor speed limit $\mathbf{n}_{\max}$ and $\mathbf{n}_{\min}$, and thus, we have the maximum joint velocity limit $\dot{\mathbf{q}}_{\max}$ and $\dot{\mathbf{q}}_{\min}$, which are obtained by multiplying the gear ratio by the motor speed limit. Therefore, the joint velocity constraint inequality equation can be expressed as:

$$\dot{\mathbf{q}}_{\min} \leq \dot{\mathbf{q}}(s) \leq \dot{\mathbf{q}}_{\max} \tag{10}$$

By substituting equation (2) into equation (10), the inequality equation is rewritten as:

$$\dot{\mathbf{q}}_{\min}/\mathbf{q}'(s) \leq \dot{s} \leq \dot{\mathbf{q}}_{\max}/\mathbf{q}'(s) \tag{11}$$

2.3. Optimization objective

The optimization goal of time-optimal path tracking is to minimize the entire trajectory execution time; thus, the objective function can be expressed as:

$$\min\ T = \int_0^T 1\,dt \tag{12}$$

By changing the integration variable from $t$ to $s$, the objective equation (14) can be rewritten as:

$$\min\ T = \int_0^T 1\,dt = \int_{s(0)}^{s(T)} \frac{1}{\dot{s}}ds = \int_0^1 \frac{1}{\dot{s}}ds \tag{13}$$

Considering the constraint conditions and the optimization objective function, the optimization problem can be expressed as:

$$\min\ T = \int_0^1 \frac{1}{\dot{s}}ds$$

$$s.t. \begin{cases} \tau_{min} \leq \mathbf{m}(s)\ddot{s} + \mathbf{c}(s)\dot{s}^2 + \mathbf{f}(s)\dot{s} + \mathbf{g}(s) \\ \tau_{max} \geq \mathbf{m}(s)\ddot{s} + \mathbf{c}(s)\dot{s}^2 + \mathbf{f}(s)\dot{s} + \mathbf{g}(s) \\ \dot{\mathbf{q}}_{min}/\mathbf{q}'(s) \leq \dot{s} \leq \dot{\mathbf{q}}_{max}/\mathbf{q}'(s) \\ \ddot{s} \leq (\ddot{\mathbf{q}}_{max} - \mathbf{q}''(s)\dot{s}^2)/\mathbf{q}'(s) \\ \ddot{s} \geq (\ddot{\mathbf{q}}_{min} - \mathbf{q}''(s)\dot{s}^2)/\mathbf{q}'(s) \\ s(0) = 0 \\ s(T) = 1 \\ \dot{s}(0) = 0 \\ \dot{s}(T) = 0 \end{cases} \tag{14}$$

**3. Q-learning algorithm and its limitations**

Q-learning is a type of reinforcement learning(RL) algorithms developed by Watkins in 1988 [26]. Q-learning applies the concept of reward and penalty in exploring an unknown environment and searching for a policy that maximizes the reward. Figure 1 shows the typical agent-environment interaction in Q-learning. In the Q-learning algorithm, the learner and decision maker are called the agents (as shown in Figure 1 [19][19]), and the objects the agent interacts with, comprising everything outside the agent, are called the environment. The agent and environment interact continually, the agent selecting actions and the environment responding to those actions and presenting new situations to the agent. The environment also gives rise to rewards, special numerical values that the agent tries to maximize over time [19]. In the time-optimal path tracking problem, the agent is a pseudo-agent that starts from the initial state (0,0) in the phase plane $s - \dot{s}$ and searches for the optimal velocity trajectory along the direction in which the pseudo-displacement increases. The environments are the constraint conditions in (9) and

(11). A state is a point $(s_k, \dot{s}_k)$ in the phase plane $s - \dot{s}$, where $k$ is the discrete point in the task path (the discrete method is described in Section 4). An action is the movement that the agent takes to move from one state to another state. A reward is a positive value given to increase the Q-value for correct actions taken by the agent at a particular state, whereas a penalty is a negative value given to decrease the Q-value for an incorrect action taken by the agent.

The learning experience of RL is obtained from the exploration and exploitation by RL agents. In general, exploration is used at the beginning of the learning process, and exploitation is used at the end of the learning process because exploration involves the selection of random actions while the agent learns without considering the current state to visit all the state-action pairs in the environment. In contrast, exploitation involves the knowledge from the agent applied in choosing actions to maximize the reward of the current state [27]. In Q-learning, it is not necessary to have complete environmental modelling such that the transition probability matrix of all states and actions is known initially. The Q-learning algorithm has been a vital contributor in the development of RL [22]. The policy being applied is not affected by the values of the optimal Q to be converged. The Q-values of the Q-learning algorithm are updated using the following expression:

$$Q(S_k, A_k) \leftarrow Q(S_k, A_k) + \alpha[R_{k+1} + \gamma \max_{A_{k+1}} Q(S_{k+1}, A_{k+1}) - Q(S_k, A_k)] \quad (15)$$

where:

$S_k$ is the current state, where in the time-optimal path tracking problem, $S_k$ is a point $(s_k, \dot{s}_k)$ in the phase plane $s - \dot{s}$;

$S_{k+1}$ is the next state, where in the time-optimal path tracking problem, $S_{k+1}$ is a point $(s_{k+1}, \dot{s}_{k+1})$ in the phase plane $s - \dot{s}$;

$A_k$ is the action performed in the $S_k$ state, where in the time-optimal path tracking problem, $A_k$ is the movement from the current state $(s_k, \dot{s}_k)$ to the next state $(s_{k+1}, \dot{s}_{k+1})$;

$A_{k+1}$ is the action performed in the $S_{k+1}$ state;

$R_{k+1}$ is the reward or penalty received from the environment when the agent takes the action $A_k$ in state $S_k$;

$\gamma$ is the discount factor $(0 \leq \gamma < 1)$;

$\alpha$ is the learning coefficient $(0 < \alpha < 1)$;

$\max_{A_{k+1}} Q(S_{k+1}, A_{k+1})$ is the maximum Q-value obtained in all possible actions of the next state $S_{k+1}$ (there is a one-to-one mapping between all the actions and the Q-table).

The pseudo-code of the classical Q-learning algorithm is summarized as in Algorithm 1.

Even though the Q-learning is able to learn and improve the solution from time to time, it encounters several Achilles' heels. Consider a searching environment with $n$ states and $m$ number of possible actions, the dimension of the constructed Q-table will be $n \times m$. When moving from the current state to the next state, the agent has to select the action with the highest Q value among the $m$ possible actions. This implies that $(m - 1)$ times of comparison is required. To update the Q-table with $n$ states, the number of comparisons needed is $n(m - 1)$. Hence, when the size and complexity of environment increase, the time took for the Q-learning to complete the path planning increases exponentially as the search space increases [28].

Furthermore, during the initial stage of exploration, the motion of agent is completely random, resulting in wastage of computational effort, slower convergence rate and time-consuming. As shown in equation (15), the action selected for the next state will be determined by the highest Q-value. The agent has no choice but to perform random selection during the early stage of learning since all Q-values are initialized to zeros. The random movement with exploring probability continues to a certain extent, up until the Q-values are refined [22].

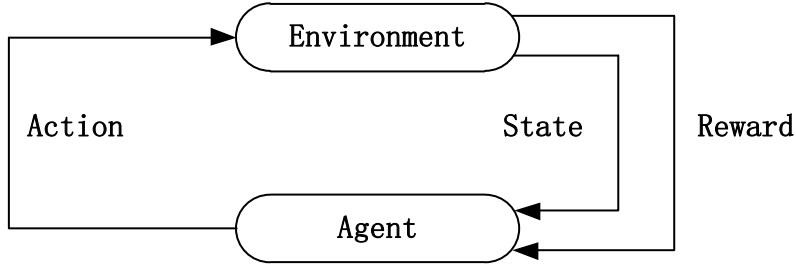

**Fig. 1**. Interaction between the Agent and the Environment

---
**Algorithm 1: Classical Q-learning algorithm**
---
   Initialize all Q-values, i.e., $Q(S, A)$ in Q-table to zero
   Repeat (for each episode):
     Initialize $S_k$
     Repeat (for each episode):
       Choose $A_k$ from $S_k$ using policy derived from Q (e.g., $\epsilon-$greedy)
       Take action $A_k$, observe $R_{k+1}$, $S_{k+1}$
       $Q(S_k, A_k) \leftarrow Q(S_k, A_k) + \alpha[R_{k+1} + \gamma \max_{A_{k+1}} Q(S_{k+1}, A_{k+1}) - Q(S_k, A_k)]$
       $S_k \leftarrow S_{k+1}$
   Until $S_k$ is terminal
---

## 4. Approaches for setting the reinforcement learning states and improving convergence rate

In order to counter the limitation as described above, some approaches for improving the Q-learning algorithm is suggested in this Section. Firstly, in order to avoid the increase of the exploration space and set the discrete reinforcement learning state, it is necessary to discretize the continuous task path of the robot manipulator and reduce the number of discrete points to avoid excessively large number of states. Secondly, due to the constraint conditions (9) (11), the pseudo-velocity in the next state should better be limited to a certain range (as the action that is out of range is certainly to break the constraints), which can reduce the times of action comparison. Finally, in order to improve the convergence rate, the initial Q value should be specialized. In this paper the optimal trajectory obtained by the direct method under the conservative torque constraints is used as the prior knowledge of the exploration process.

### 4.1. Setting the discrete reinforcement state
### 4.11. Path discretization

To construct the time-optimal path tracking problem as an RL problem, it is necessary to set the discrete state points, which requires discretizing the continuous path and assuming that the motion between any adjacent discrete points is uniformly accelerated motion. This treatment method was discussed in [9].

From equations (4), (5), (6), (7) and (8), we know that the values of $\boldsymbol{\tau}(s)$, $\mathbf{m}(s)$, $\mathbf{c}(s)$, $\mathbf{f}(s)$ and $\mathbf{g}(s)$ are related to the pseudo-curvature $\mathbf{q}'(s)$ and the change rate of the pseudo-curvature $\mathbf{q}''(s)$, where $\mathbf{q}'(s)$ and $\mathbf{q}''(s)$ are nonlinear functions about the pseudo-displacement $s$. Therefore, the maximum acceleration between two discrete points obtained by equation (9) is also a nonlinear function of pseudo-displacement $s$ and pseudo-velocity $\dot{s}$; i.e., the motion between two adjacent discrete points is actually a variable acceleration motion. Therefore, the approximation treatment that regards the motion between two adjacent points as uniformly accelerated motion should be cautious, as a sudden change in $\mathbf{q}'(s)$ and $\mathbf{q}''(s)$ between two discrete points may lead to greatly overrun torque and cause abnormal conditions or even shut down during the trajectory execution. To prevent the occurrence of greatly overrun torque, the discrete points should be selected by some rule to avoid the sudden change in $\mathbf{q}'(s)$ and $\mathbf{q}''(s)$ between any adjacent points. Setting thresholds ε and σ to control the pseudo-curvature difference $\Delta \mathbf{q}'(s)$ and the change rate difference $\Delta \mathbf{q}''(s)$ between arbitrary adjacent points to a certain range prevents the calculated torque from greatly exceeding the constraint limit. The workflow of this selective discrete method is shown in Figure 2. Through this method, the situation in which the calculated torque greatly exceeds the torque constraints is avoided, and the discrete points used for RL are decreased, which helps reduce the state space and computation time.

The task path is discretized into N points by the method shown in Figure 2; thus, the optimization problem can be discretized as:

$$\min \sum_{1}^{N} \frac{1}{\dot{s}_k}$$

$$s.t. \begin{cases} \underline{\boldsymbol{\tau}}(s_k, \dot{s}_k) \leq \mathbf{m}(s_k)\ddot{s}_k + \mathbf{c}(s_k)\dot{s}_k^2 + \mathbf{f}(s_k)\dot{s}_k + \mathbf{g}(s_k) \leq \overline{\boldsymbol{\tau}}(s_k, \dot{s}_k) \\ \underline{\dot{\mathbf{q}}}(s_k)/\mathbf{q}'(s_k) \leq \dot{s}_k \leq \overline{\dot{\mathbf{q}}}(s_k)/\mathbf{q}'(s_k) \\ (\underline{\ddot{\mathbf{q}}}(s_k) - \mathbf{q}''(s_k)\dot{s}_k^2)/\mathbf{q}'(s_k) \leq \ddot{s}_k \leq (\overline{\ddot{\mathbf{q}}}(s_k) - \mathbf{q}''(s_k)\dot{s}_k^2)/\mathbf{q}'(s_k) \\ s_1 = 0 \\ s_N = 1 \\ \dot{s}_1 = 0 \\ \dot{s}_N = 0 \end{cases} \quad (16)$$

$$\text{for } k = 1 \cdots N.$$

where, for convenience, the upper and lower limits are indicated by the upper subscript $\overline{\phantom{x}}$ and lower subscript $\underline{\phantom{x}}$, respectively.

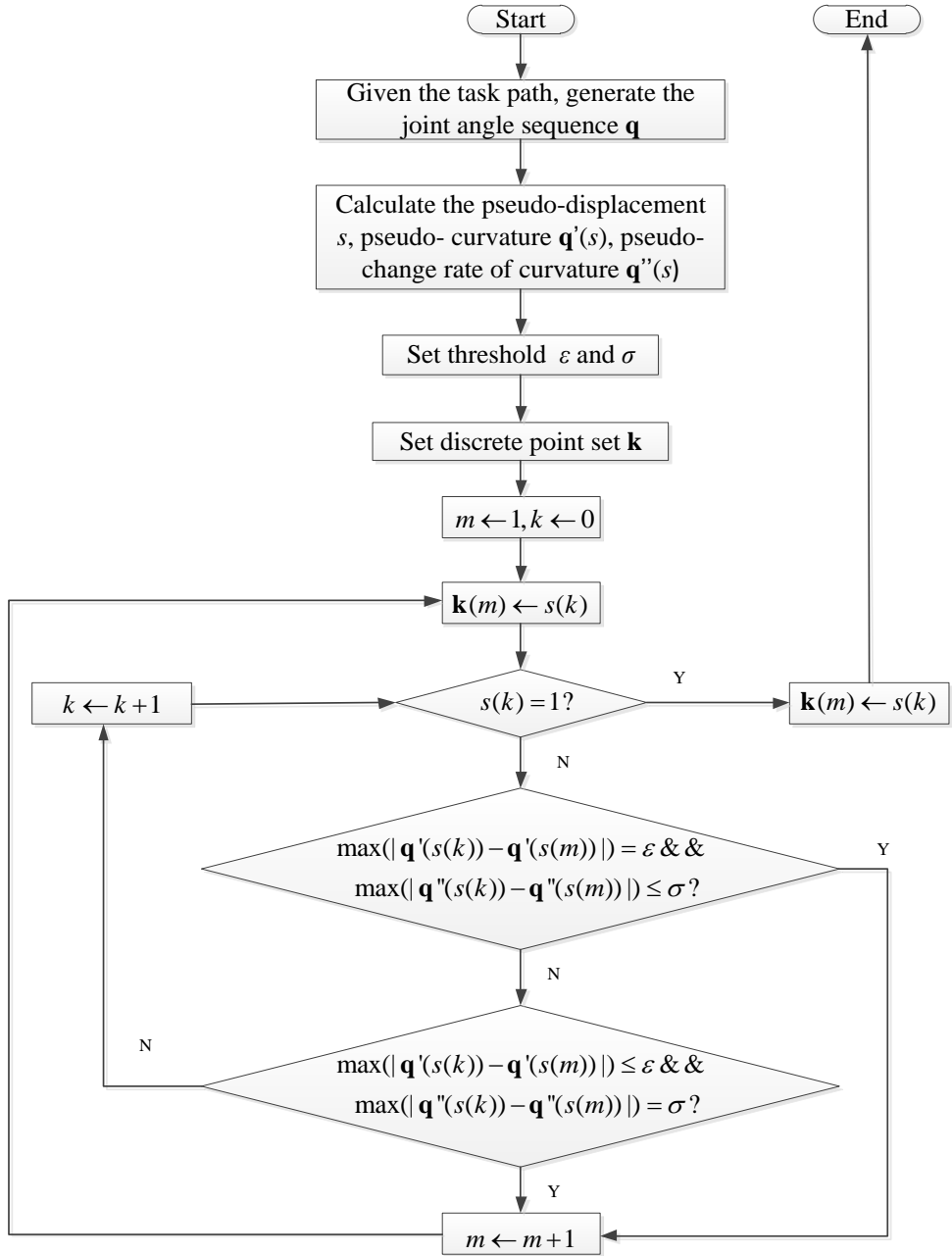

**Fig. 2**. Working flow of the path discretization

4.12 Divide the phase plane $s - \dot{s}$ into a grid

Since the task path has been discretized into $N$ points, i.e., the $s$-coordinate direction is divided into $N$ grid points, dividing the $\dot{s}$-coordinate into $M$ grid points is considered. As the maximum pseudo-velocity $\max(\dot{\mathbf{q}}_{max}/\mathbf{q}'(s))$ can be obtained according to equation (11), taking $\max(\dot{\mathbf{q}}_{max}/\mathbf{q}'(s))/M$ as the grid length to divide the $\dot{s}$ coordinate into $M$ grid points yields an $N \times M$ grid.

4.2 Calculate the range of action

Since the motion between adjacent discrete points is assumed to be uniformly accelerated motion, the maximum and minimum pseudo-velocity values can be calculated according to the uniformly accelerated motion equations; thus, the RL actions (i.e., the pseudo-velocity) between the maximum and minimum pseudo-velocities are feasible actions, and the other actions are unfeasible actions (which break the constraint conditions). The uniformly accelerated motion equations are expressed as follows:

$$\dot{s}_{k+1}^2 - \dot{s}_k^2 = 2\ddot{s}_k(s_{k+1} - s_k) \tag{17}$$

$$\dot{s}_{k+1} = \sqrt{2\ddot{s}_k(s_{k+1} - s_k) + \dot{s}_k^2} \tag{18}$$

where $\dot{s}_k$ is the pseudo-velocity of the current discrete point, $\dot{s}_{k+1}$ is the pseudo-velocity of the next discrete point and $\ddot{s}_k$ is the pseudo-acceleration of the current discrete point.

The maximum and minimum pseudo-acceleration can be calculated according to equation (9); thus, we have

$$\ddot{s}_{max,k} = \min((\boldsymbol{\tau}_{\max,k}(s_k,\dot{s}_k) - \mathbf{c}(s_k)\dot{s}_k^2 - \mathbf{f}(s_k)\dot{s}_k - \mathbf{g}(s_k)/\mathbf{m}(s_k)) \tag{19}$$

$$\ddot{s}_{min,k} = \max((\boldsymbol{\tau}_{\min,k}(s_k,\dot{s}_k) - \mathbf{c}(s_k)\dot{s}_k^2 - \mathbf{f}(s_k)\dot{s}_k - \mathbf{g}(s_k)/\mathbf{m}(s_k)) \tag{20}$$

Substituting equations (20) and (21) into equation (19) yields the maximum and minimum pseudo-velocities for the next discrete point as follows:

$$\dot{s}_{max,k+1} = \sqrt{2\ddot{s}_{max,k}(s_{k+1} - s_k) + \dot{s}_k^2} \tag{21}$$

$$\dot{s}_{min,k+1} = \sqrt{2\ddot{s}_{min,k}(s_{k+1} - s_k) + \dot{s}_k^2} \tag{22}$$

Then, the action range of the next discrete point is limited to $[\dot{s}_{min,k+1},\dot{s}_{max,k+1}]$, which avoids other unnecessary searches.

4.4 Acquire prior knowledge

In order to avoid a large number of computations, it is necessary to use the prior knowledge before the learning process is starting so as to specify initial Q values, which is help for improving convergence rate. The method of improving the reinforcement learning convergence rate through prior knowledge has been applied in [18, 24]. In this paper, the optimal trajectory obtained by the direct method under the conservative torque constraints is used as the prior knowledge of the exploration process.

As most of the direct methods for time-optimal path tracking problems are calculated by numerical integration, while they are not suitable for the case of the discretized path in this paper. In [29], a numerical integration-like (NI-like) time-optimal path tracking approaches for a discretized path is proposed, which is planned by using uniform acceleration equations instead of numerical integration. However, since the method of [29] does not divide the phase plane $s - \dot{s}$ into a grid, the planning method needs to be modified to make it suitable for the case of grid mode. The modifications are shown as follow:

Modify the planning method of computing forward: In [29], the pseudo-velocity of the next discrete point $\dot{s}_{k+1}$ is calculated from the pseudo-velocity of the current discrete point $\dot{s}_k$ by the uniform acceleration equation with the maximum pseudo-acceleration. However, since the phase plane $s - \dot{s}$ is divided into an $N \times M$ grid, the pseudo-velocity of the next discrete point should be the pseudo-velocity of the grid point which closest to $\dot{s}_{k+1}$ and not greater than $\dot{s}_{k+1}$, denoted by $\dot{s}'_{k+1}$ (symbol ' for distinguish from the pseudo-velocity $\dot{s}_{k+1}$.)

Modify the planning method of computing backward: Similar to the improvement of computing forward, the pseudo-velocity of the previous discrete point should be the pseudo-velocity of the grid point which closest to the calculated pseudo-velocity of the previous points $\dot{s}_{k-1}$ and not greater than $\dot{s}_{k-1}$, denoted by $\dot{s}'_{k-1}$.

For distinguish, we named the above modified methods as NI-like grid mode (NIGM) in this paper.

5. Reinforcement learning algorithms for robotic time-optimal path tracking

5.1 Improved Q-learning (IQL) algorithm for robotic time-optimal path tracking

Considering the Q-learning algorithm and its limitations, as mentioned in Section 3, the Q-learning algorithm is improved to make it more suitable for solving the time-optimal path tracking problem. Combined with the improved approaches in Section 4, the steps of the IQL algorithm are as follows:

Step 1. Discretize the task path into *N* points using the method in Section 4.11;

Step 2. Divide the phase plane $s - \dot{s}$ into an $N \times M$ grid using the method in Section 4.12;

Step 3 Calculate the optimal trajectory under conservative torque constraints using the method in Section 4.3 as the prior knowledge, and then substitute the obtained trajectory into equations (20) and (21) to determine whether a state in the trajectory violates the velocity-dependent torque constraints (if the minimum pseudo-acceleration obtained by equation (20) at a state is greater than the maximum pseudo-acceleration obtained by equation (21), then this state breaks the velocity-dependent torque constraints). Add a positive value to the Q-table corresponding to the states that do not violate constraints and add a negative value in the Q-table corresponding to the states that violate constraints. In addition, the last part of the optimal trajectory that does not violate constraints is used as a terminate state, and the episode ends when the agent reaches or crosses one of the terminate states. Figure 3 shows typical prior knowledge in the phase plane $s - \dot{s}$;

Step 4. Set (0,0) as the initial state;

Step 5. Start from the initial state to explore and obtain the action in the next state. The exploration method uses the ε-greedy algorithm to select an action from the action range, which is calculated by the method proposed in Section 4.2, where to improve the convergence rate, the actions corresponding to negative Q-values in the calculated action range should be neglected, as these actions will direct the agent to the states that violate constraints);

Step 6. Determine whether the agent reaches or crosses one of the terminate states, i.e., the trajectory obtained by the agent's exploration intersects with the segment combined with the terminate states. If the exploration intersects, rewarding the selected action, then update the corresponding Q-value by the action-value function equation (15), end this episode, and go to Step 8 for exploitation. If the exploration does not intersect, then go to Step 7 to determine whether the next state that the agent reaches violates the constraint conditions;

Step 7. Determine whether the next state that the agent reaches violates the constraint conditions. If the state does not violate the constraint conditions, then reward the selected action, update the corresponding Q-value by equation (15), set the next state as the initial state, and return to Step 5 to continue the learning process. If the state violates the constraint condition, then penalize the selected action, update the corresponding Q-value by equation (15), and return to Step 4 to restart the learning process. (The determination for violating the constraint is similar to that in Step 3. In addition, if the Q-value corresponding to all possible actions of the next state is negative, it is also considered a violation of the constraints, as these actions will direct the agent to the states that violate constraints.)

Step 8. Let $\varepsilon = 0$ exploit the RL experience, obtain the optimal trajectory and the corresponding return;

Step 9. Repeat Steps 4-8 until the return is no longer updated or the total number of episodes is greater than the set number of maximum episodes.

From the optimization objective equation (17), we know that to obtain an optimal trajectory, the pseudo-velocity of the planned trajectory should be as large as possible, which requires a greater reward for the actions that can obtain a larger pseudo-velocity. Therefore, the RL reward and penalty are designed to be associated with the pseudo-velocity as follows:

$$R_{k+1} = \begin{cases} \dot{s}_k + \dot{s}_{k+1} & reward \\ -\mu(\dot{s}_k + \dot{s}_{k+1}) & penalty \end{cases} \quad (23)$$

where μ is a penalty factor that is used to increase the penalty for actions that violate the constraint.

In RL, return is a specific function used to define the agent's goal, i.e., to finally maximize the cumulative reward. In the simplest case, the return is the sum of the rewards [19]. Since the optimization objective is to maximize the pseudo-velocity in the time-optimal path tracking problem, the return $G_t$ is defined as the sum of the pseudo-velocity in all discrete states:

$$G_t = \sum_{k=1}^{N} \dot{s}_k \quad (24)$$

The workflow of the proposed IQL algorithm for robotic time-optimal path tracking is shown in Figure 4.

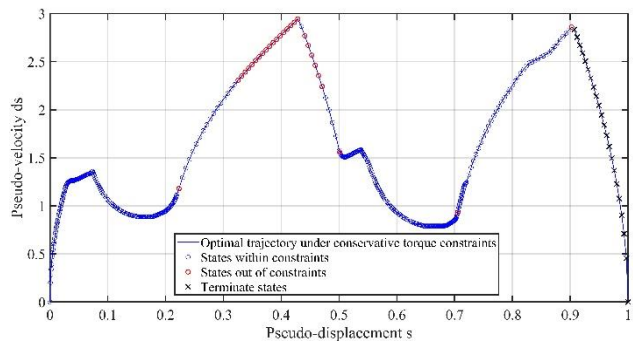

Figure 3. A typical prior knowledge in the phase plane $s - \dot{s}$

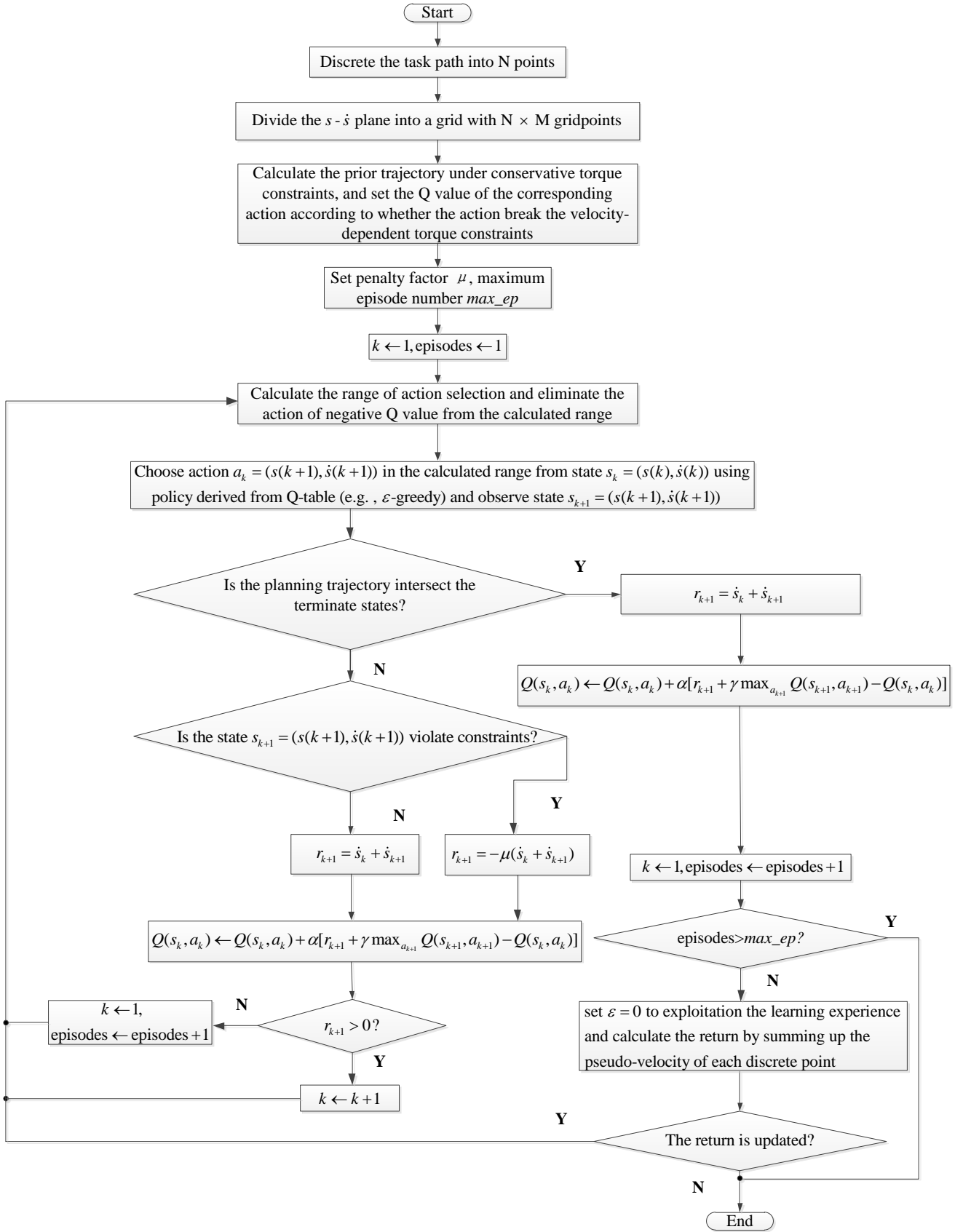

Figure 4. Workflow of the proposed IQL algorithm for robotic time-optimal path tracking

5.2 Improved action-value function RL (IAVRL) algorithm for robotic time-optimal path tracking

Although the IQL algorithm can improve the limitations of the Q-learning algorithm, there are still some problems in solving the time-optimal path tracking problems. In the Q-learning algorithm, the Q-value corresponding to an action is related to the number of times that the action has been chosen. The action corresponding to the maximum Q-value is selected with the highest probability under the ε-greedy policy, and the Q-value increases after the action is selected by the agent interacting with the environment within the limits of the constraints, which creates a loop. In the case of Q-table initialization, when one of several adjacent actions of a state is selected by the ε-greedy policy, the selected action has a larger Q-value than other actions. A larger Q-value increases the likelihood of an action being selected in the next episode, which may result in an increase in the number of times the action is selected. Additionally, an increase in the number of times the action is selected will increase the corresponding Q-value, which increases the likelihood of the action being selected in the next episode. The above loop will cause the agent to become stuck in the local optimal solution and require many exploration episodes to obtain the optimal solution. In addition, for the time-optimal path tracking problem, as with many numerical integration methods, before the planning trajectory intersecting the model view controller (MVC), the acceleration for the planning trajectory should be changed from maximum acceleration to minimum acceleration to avoid intersecting the MVC. However, since Q-learning is single-step updated, the Q-value corresponding to the current state is related to only the reward of the current state and the maximum Q-value of the next state. Therefore, using the action-value function (2) to update the Q-table requires considerable exploration time.

To improve the above-mentioned problems, a new action-value function is proposed as follows:

$$Q(S_k, A_k) \leftarrow R_{k+1} + \rho^{K-k} R_{K+1} \qquad (25)$$

where:

$S_k$ is the current state;

$A_k$ is the action performed in $S_k$ state;

$R_{k+1}$ is the reward received from the environment when the agent takes the action $A_k$ in state $S_k$;

$K$ refers to the step in which the agent takes action $A_K$ in the state of $S_K$ and violates the constraints;

$R_{K+1}$ is the penalty received from the environment when the agent takes the action $A_K$ in state $S_K$;

$\rho$ is the discount factor $(0 < \rho < 1)$. By using the discount factor, the actions in the states that are closer to the constraint boundary will receive a greater penalty.

By using the improved action-value function (25), the episodes that do not violate the constraints receive a greater return than the episodes that violate the constraints. Additionally, in all the episodes that do not violate the constraints, the episodes that have greater pseudo-velocity receive greater rewards and thus obtain a greater return. Therefore, the time-optimal path tracking trajectory is the episode that does not violate the constraints and has the greatest pseudo-velocity in every state of the episode. Moreover, the action-value function (25) is a multi-step update function. When the agent takes action $A_K$ in state $S_K$ and violates the constraints, all the Q-values in this episode can be updated at once by equation (25), which improves the convergence rate.

Combined with the improved approaches in Section 4, the steps of the improved action-value function RL algorithm are similar to the steps proposed in Section 5.1, except that the update of the Q-table and the exploration in the RL. Different from the Q-learning algorithm, the IAVRL algorithm performs exploration only when there are actions in the action range that are not selected. When all the actions in the action range have been selected, the algorithm no longer performs exploration. The workflow of the IAVRL algorithm for robotic time-optimal path tracking is shown in Figure 4.

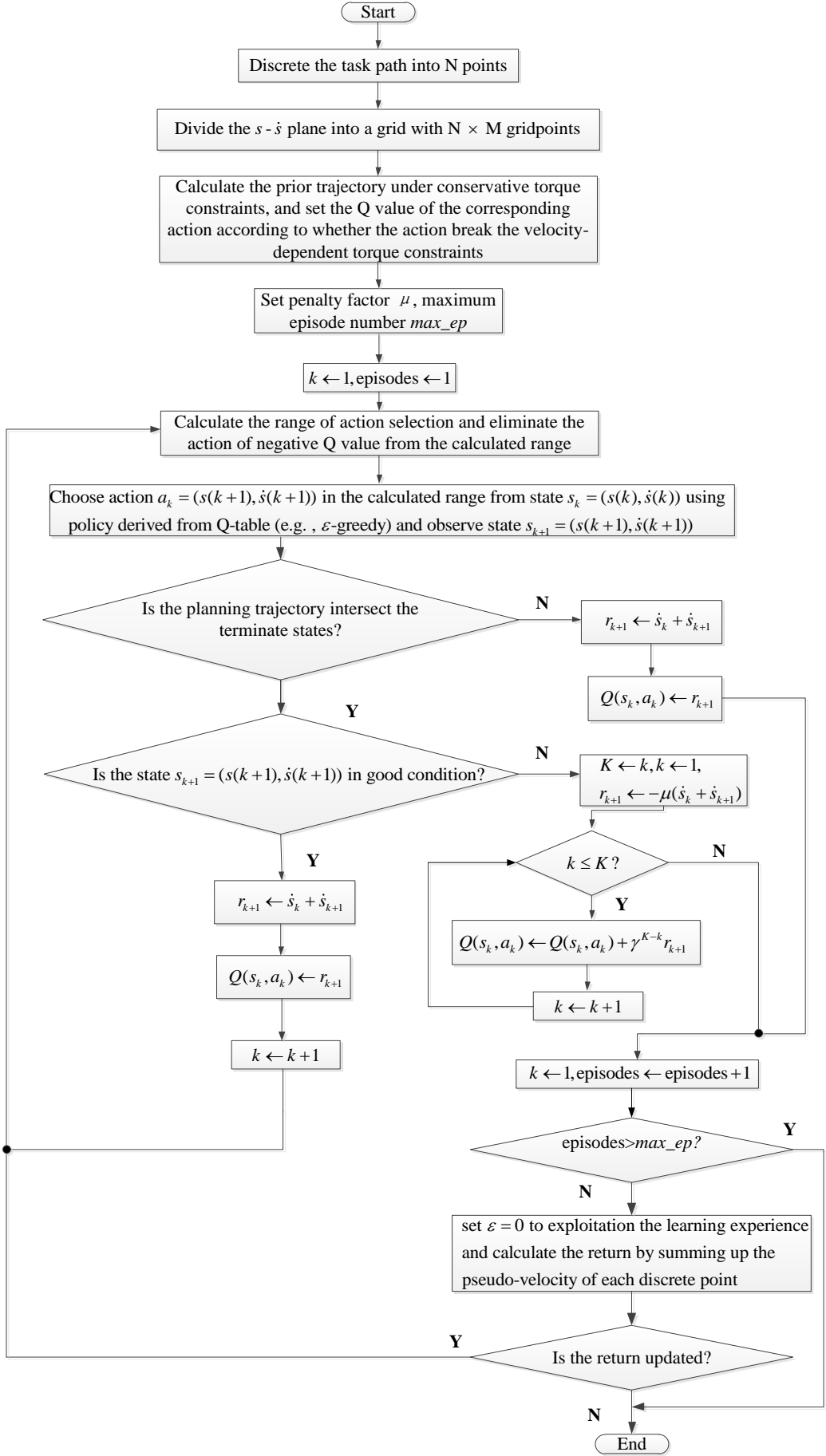

Figure 5. Working flow of the proposed IAVRL algorithm for robotic time-optimal path tracking

## 6. Experiment results and performance analysis

6.1 Experimental settings

*Configuration environment for implementation*

All the RL algorithms are implemented in MATLAB R2018b on an Intel Core i7 CPU running at 3.40 GHz on a Windows machine.

*Industrial robot for experiment*

The industrial robot used for experimental verification is a 6-DOF GSK-RB03A1 robot of Guangzhou CNC Equipment Co., Ltd, as shown in Figure 6. The dynamic model and dynamic parameter of the robot are obtained by the method of [25, 30]. The servo motor torque characteristics of each joint of the robot are shown in Figure 7.

*Task path*

The task path is given in the Cartesian space as a rounded rectangle with a diameter of 100 mm and a centre distance of 200 mm, as shown in Figure 8.

*Threshold for path discretization*

In order to avoid the occurrence of greatly overrun torque during the trajectory execution, the threshold $\varepsilon$ to control the pseudo-curvature difference $\Delta \mathbf{q}'(s)$ is set to 0.01 and the threshold $\sigma$ to control the change rate difference $\Delta \mathbf{q}''(s)$ is set to 0.1. Therefore, the task path is discretized into 527 points.

*The division of grids*

The phase plane is divided into grids of $527 \times 500$、$527 \times 1000$、$527 \times 1500$、$527 \times 2000$, respectively.

*Reinforcement learning parameters*

The discount factor $\gamma$ for IQL is set to 0.8, the learning coefficient $\alpha$ for IQL is set to 0.8. The discount factor $\rho$ for IAVRL is set to 0.8. The penalty factor $\mu$ for the penalty of two algorithms is set to 1.25. The maximum episode numbers is set to 500,000. Greed factor of $\epsilon - greedy$ is set to 0.4.

*Q values of prior knowledge*

The Q values of prior knowledge for IQL is set as:

$$Q(S_k, A_k) = \begin{cases} 25(\dot{s}_k + \dot{s}_{k+1}), & within\ constraints \\ -25(\dot{s}_k + \dot{s}_{k+1}), & out\ of\ constraints \end{cases}$$

The Q values of prior knowledge for IAVRL is set as:

$$Q(S_k, A_k) = \begin{cases} \dot{s}_k + \dot{s}_{k+1}, & within\ constraints \\ -1.25(\dot{s}_k + \dot{s}_{k+1}), & out\ of\ constraints \end{cases}$$

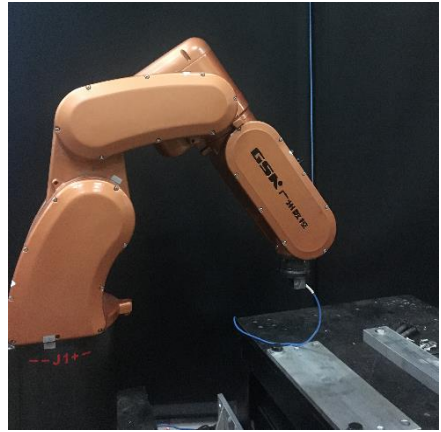

Figure 6. 6-DOF GSK-RB03A1 robot

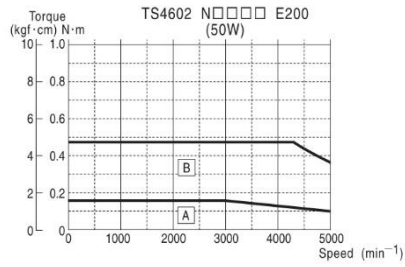

(a)

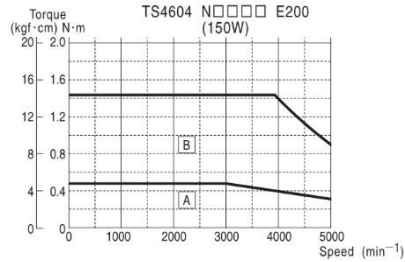

(b)

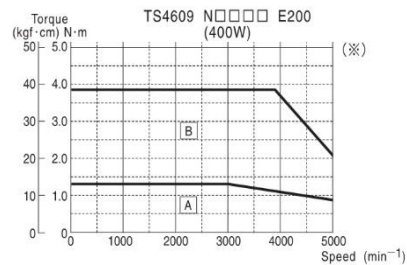

(c)

Figure 7. (a) Servo motor torque characteristics of joints 4, 5, and 6; (b) Servo motor torque characteristics of joint 3; (c) Servo motor torque characteristics of joints 1 and 2. (Ⓐ is the continuous operation area; Ⓑ is the acceleration/deceleration area)

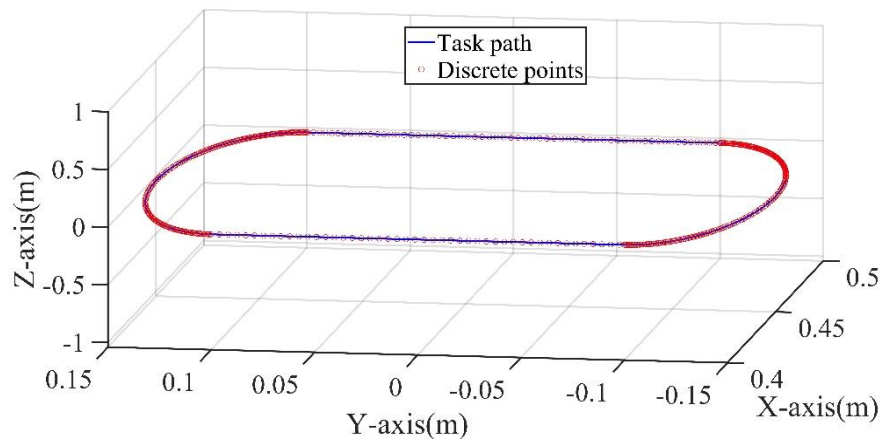

Figure 8. Task path in Cartesian space

6.2 Experiment results and analysis

6.2.1 Comparison experiment regarding path discretization

To verify the effectiveness and necessity of the selective discrete method proposed in Section 4.1, the uniform discrete method that uniformly discretizes the task path into 527 points is chosen as the comparison method. To eliminate the influence of unrelated variables, the two discrete methods are both combined with the NI-like method of [29] to plan the optimal trajectory and obtain the corresponding calculated torque. The calculated torques of the optimal trajectories, which are obtained based on the two discrete methods, are shown in Figure 9. The results show that although the calculated torques obtained based on the selective discrete method were not all within the torque constraints, they did not greatly exceed the torque constraints, verifying the effectiveness of the method. However, the calculated torques obtained based on the uniform discrete method greatly exceeded the torque constraints in joint 2 and joint 5, verifying the necessity of the selective discrete method as the calculated torques obtained based on the selective discrete method did not greatly exceed the torque constraints.

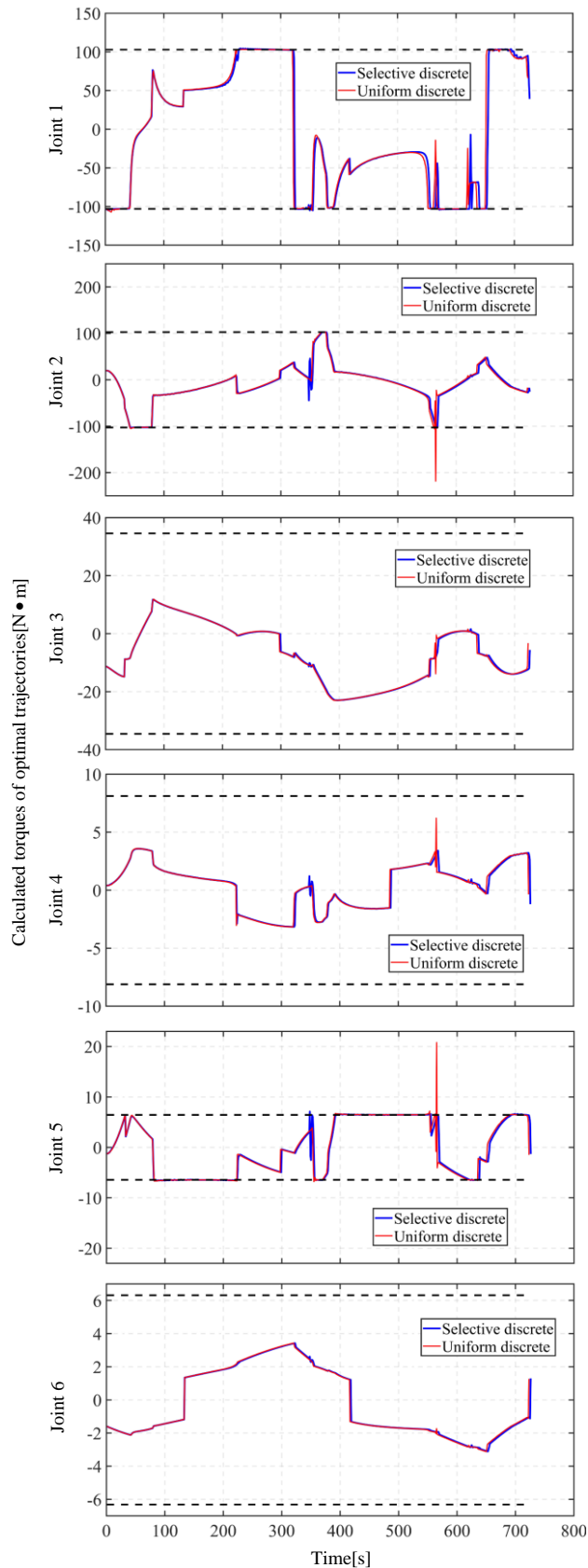

Figure 9. Calculated torques of the optimal trajectories that were obtained based on the two discrete methods

6.2.2 Comparison experiment regarding the RL algorithm under conservative constraints

To verify the feasibility of the RL algorithm in solving the time-optimal path tracking problem, this paper first considers implementing RL algorithms under conservative constraints. The experiment workflows are similar to the methods shown in Figure 4 and Figure 5 except that the workflows do not consider prior knowledge and the torque constraints are conservative. The direct planning NI-like method in [29] is used as the comparison algorithm. Table 1 shows the performance comparison among the proposed IQL algorithm, IAVRL algorithm, the NI-like algorithm in [29] and the NIGM algorithm (which is a modified version of the NI-like algorithm proposed in Section 4.4), where to reduce the experimental error, all the results shown in Table 1 are averages of 10 calculations. Table 2 shows the performance percentage by IQL and IAVRL compared with that of the NI-like and NIGM algorithms. Figure 10 shows the planned results in the phase plane $s - \dot{s}$ when the four algorithms converge or the maximum number of episodes is reached. Figure 11 shows the return obtained by the IQL and IAVRL algorithms by exploiting the learning experiences after a successful exploration, where for better comparison, the IQL and IAVRL all implement 500,000 episodes.

From the results in Table 1 and Table 2, it can be seen that although the IQL and IAVRL algorithms cannot fully achieve the performance of the directly planned NI-like and NIGM methods and take more computational time to obtain the optimal trajectory, the obtained optimal trajectories are very close to the optimal solution, and as the dimensions of the grid increase, the obtained optimal trajectories are closer to the optimal solution. The reason for the difference between the optimal trajectory obtained by the IQL, IAVRL and NIGM and that obtained by NI-like is that the optimal pseudo-velocity is approximated to the pseudo-velocity of a grid point close to it in phase plane $s - \dot{s}$. From the results of Table 1, Table 2 and Figure 10, the feasibility of the proposed RL algorithm is verified.

In addition, from the results in Table 1, Table 2 and Figure 11, it can be seen that the proposed improved action-value function of IAVRL is more suitable for the time-optimal path tracking problem than the action-value function of IQL, as the performance and convergence rate of IAVRL are all better than those of IQL.

Table 1 Performance comparison among IQL, IAVRL, NI-like and NIGM

| Grid | Algorithm | First successful episode[1] | Converge?[2] | Convergence episode[3] | Computation time (s)[4] | Return[5] | Execution time (s)[6] |
|---|---|---|---|---|---|---|---|
| 527 × 500 | NI-like | — | — | — | 1.483 | 684.3996 | 0.7245 |
| | NIGM | — | — | — | 2.431 | 646.6633 | 0.7631 |
| | IQL | 8684.8 | No | — | 7300.313 | 637.7956 | 0.7753 |
| | IAVRL | 7037.3 | Yes | 7436.5 | 20.177 | 637.2946 | 0.7773 |
| 527 × 1000 | NI-like | — | — | — | 1.483 | 684.3996 | 0.7245 |
| | NIGM | — | — | — | 2.582 | 664.6146 | 0.7460 |
| | IQL | 18113.6 | No | — | 7331.071 | 654.2342 | 0.7589 |
| | IAVRL | 13334.1 | Yes | 13981.1 | 37.035 | 659.9750 | 0.7506 |
| 527 × 1500 | NI-like | — | — | — | 1.483 | 684.3996 | 0.7245 |
| | NIGM | — | — | — | 2.467 | 670.8873 | 0.7399 |
| | IQL | 27235.2 | No | — | 7310.520 | 660.0567 | 0.7480 |
| | IAVRL | 19675.3 | Yes | 20608.3 | 55.231 | 668.0821 | 0.7435 |
| 527 × 2000 | NI-like | — | — | — | 1.483 | 684.3996 | 0.7245 |
| | NIGM | — | — | — | 2.630 | 674.0195 | 0.7365 |
| | IQL | 36280.5 | No | — | 7343.368 | 663.0715 | 0.7450 |
| | IAVRL | 24962.0 | Yes | 26084.6 | 72.640 | 672.2361 | 0.7387 |

1. First successful episode is the episode number in which the agent first reaches or crosses one of the terminal states;
2. The determination of convergence is whether the algorithm converges before reaching the maximum number of episodes;
3. A convergence episode is the episode number in which the algorithm converges;
4. Computation time is the time from the start to the end of the program;
5. Return is the return of the last episode.
6. Execution time is the execution time of the optimal trajectory obtained from the last episode.

Table 2 Performance percentage by IQL, IAVRL compared with NI-like, NIGM

| Grid | Algorithm | Performance percentage of return compared with NI-like and NIGM (%) | | Performance percentage of optimal trajectory execution time compared with NI-like and NIGM (%) | |
|---|---|---|---|---|---|
| | | NI-like | NIGM | NI-like | NIGM |
| 527 × 500 | IQL | 93.2 | 98.6 | 107.0 | 101.6 |
| | IAVRL | 93.1 | 98.6 | 107.3 | 101.9 |
| 527 × 1000 | IQL | 95.6 | 98.4 | 104.7 | 101.7 |
| | IAVRL | 96.4 | 99.3 | 103.6 | 100.6 |
| 527 × 1500 | IQL | 96.4 | 98.4 | 103.2 | 101.1 |
| | IAVRL | 97.6 | 99.6 | 102.6 | 100.5 |
| 527 × 2000 | IQL | 96.9 | 98.4 | 102.8 | 101.1 |
| | IAVRL | 98.2 | 99.7 | 101.9 | 100.3 |

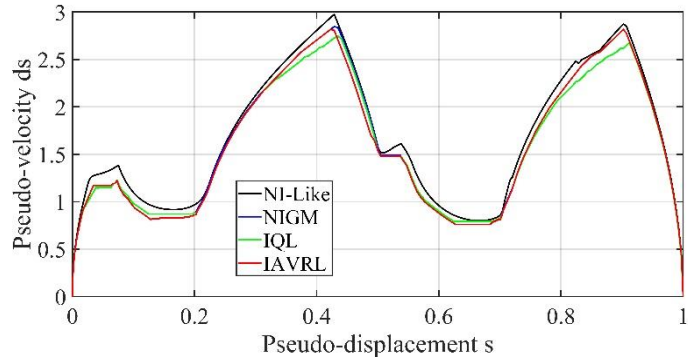

(a)

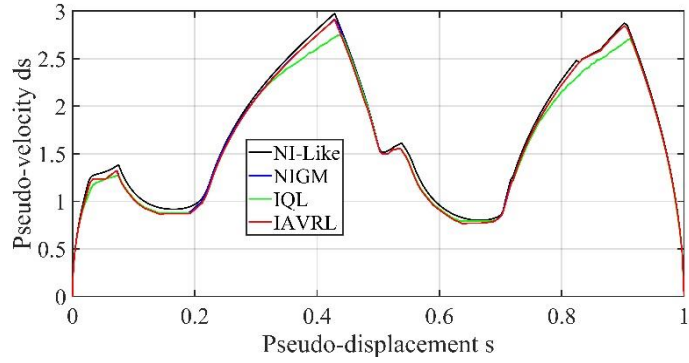

(b)

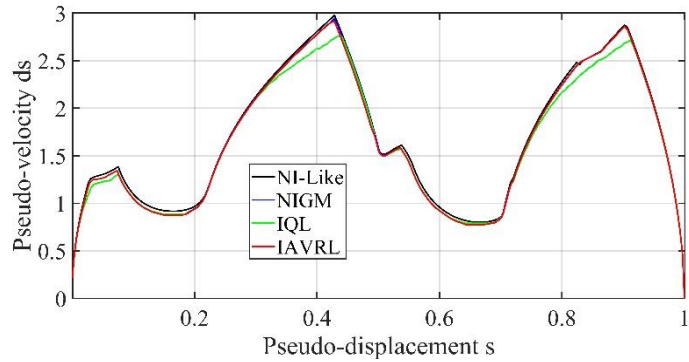

(c)

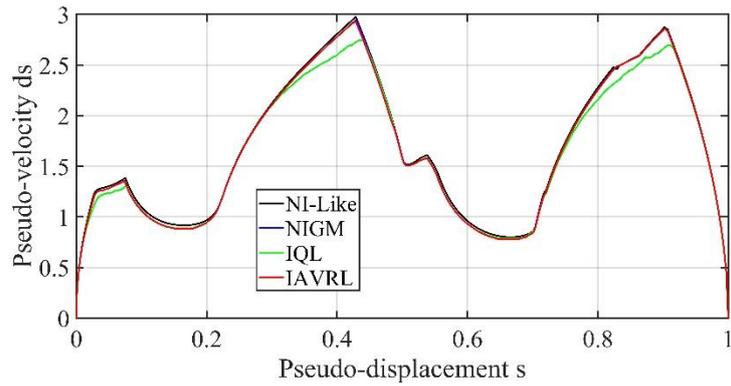

(d)

Figure 10 Planned results in the phase plane $s-\dot{s}$ when the four algorithms converge or the episodes reach the maximum number of episodes, where (a) is the case of $527\times 500$ grid, (b) is the case of $527\times 1000$ grid, (c) is the case of $527\times 1500$ grid and (d) is the case of $527\times 2000$ grid

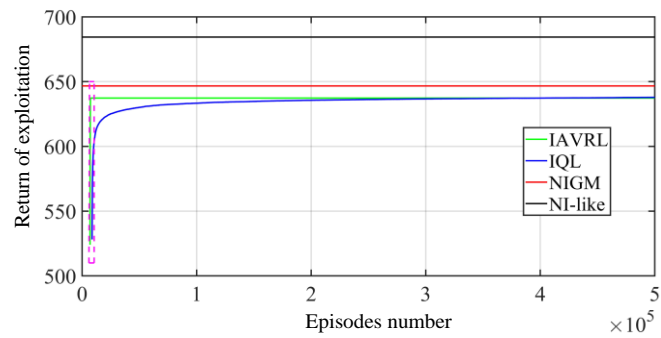

(a)

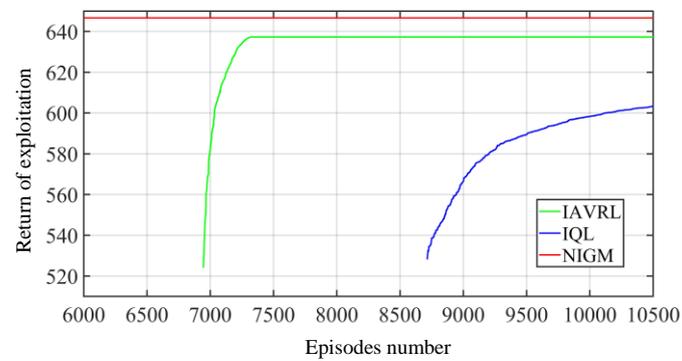

(b)

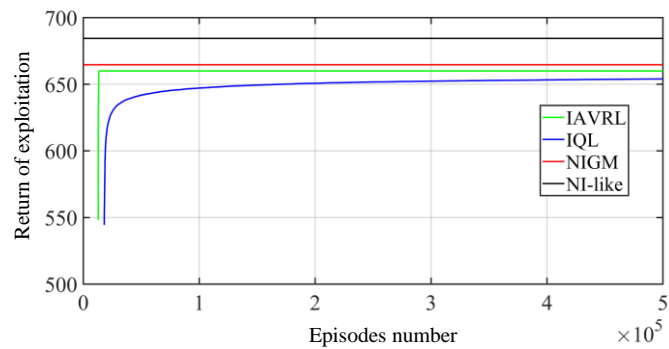

(c)

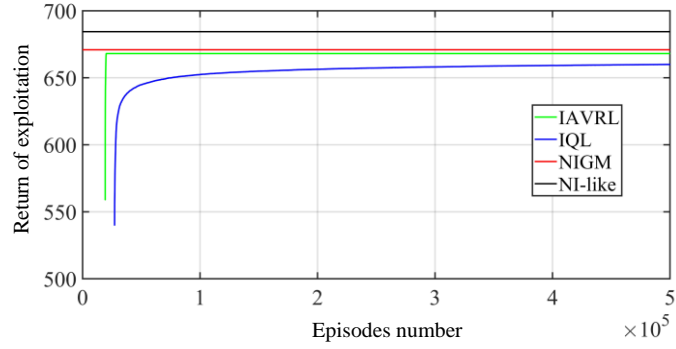

(d)

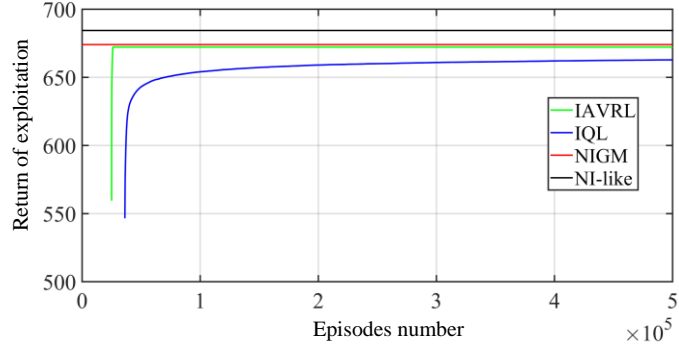

(e)

Figure 11. Return obtained by IQL and IAVRL by exploiting the learning experience after a successful exploration, where (a) is the case of a 527×500 grid, (b) is the close-up view (zoomed-in view of the pink box in (a)) to show the detailed convergence of IAVRL, (c) is the case of a 527×1000 grid, (d) is the case of a 527×1500 grid and (e) is the case of a 527×2000 grid.

6.2.2 Comparison experiment regarding RL algorithms under velocity-dependent torque constraints

The effectiveness of the proposed algorithms in solving time-optimal path tracking problems under conservative constraints is verified in Section 6.2.2. In this section, we verify the effectiveness of the proposed RL algorithms under velocity-dependent torque constraints. The experimental workflows are shown in Figure 4 and Figure 5.

Table 3 shows the performance comparison between the proposed IQL algorithm and the IAVRL algorithm, where for verifying the effectiveness of prior knowledge, the IQL algorithm without using prior knowledge and the IAVRL algorithm without using prior knowledge are set as the comparison algorithms. In addition, to reduce the experimental error, all the results shown in Table 3 are the average of 10 calculations. Table 4 shows the performance improvement achieved by using prior knowledge compared with the case without using prior knowledge. Figure 12 shows the return obtained by IQL with prior knowledge and IQL without prior knowledge by exploiting the learning experience after a successful exploration. Figure 13 shows the return obtained by IAVRL with prior knowledge and IAVRL without prior knowledge by exploiting the learning experience after a successful exploration. From the results in Table 3, Table 4, Figure 12 and Figure 13, it can be seen that by using prior knowledge, for the IQL algorithm, the convergence rate and RL return can be improved, and the computation time and optimal trajectory execution time can be reduced. For the IAVRL algorithm, the convergence rate can be improved, and the computation time can be reduced, while the convergence rate and optimal trajectory execution time did not improve. Figure 14 shows the calculated torque of the optimal trajectory obtained by IAVRL with prior knowledge in a 527×2000 grid case. From the results in Figure 14, the effectiveness of the proposed algorithms in solving the time-optimal path tracking problem under velocity-dependent torque constraints is verified, as the calculated torques do not exceed the velocity-dependent torque constraints (this is similar to the other cases that are not shown). The RL process and trajectory execution experiment can be seen at https://youtu.be/k1rg6IL4OsE.

Table 3 Performance comparison between IQL and IAVRL under velocity-dependent torque constraints

| Grid | Algorithm | Use prior knowledge? | First successful episode | Converge? | Convergence episode | Computation time (s) | Return | Execution time (s) |
|---|---|---|---|---|---|---|---|---|
| 527 × 500 | IQL | Yes | 8480.3 | Yes | 112834.5 | 2367.051 | 642.6754 | 0.7690 |
| | | No | 8466.5 | No | — | 10528.942 | 637.0641 | 0.7716 |
| | IAVRL | Yes | 6703.2 | Yes | 6772.3 | 19.894 | 633.7976 | 0.7805 |
| | | No | 9298.7 | Yes | 9826.4 | 32.010 | 646.8056 | 0.7680 |
| 527 × 1000 | IQL | Yes | 17611.1 | No | — | 10507.657 | 656.6216 | 0.7536 |
| | | No | 17596.9 | No | — | 10513.355 | 653.5335 | 0.7557 |
| | IAVRL | Yes | 12178.2 | Yes | 12541.5 | 39.222 | 657.6727 | 0.7553 |
| | | No | 17129.5 | Yes | 18032.0 | 57.957 | 664.8388 | 0.7474 |
| 527 × 1500 | IQL | Yes | 26482.3 | No | — | 10513.092 | 663.7358 | 0.7462 |
| | | No | 26482.3 | No | — | 10533.373 | 659.3529 | 0.7487 |
| | IAVRL | Yes | 17985.6 | Yes | 18890.7 | 59.854 | 666.3909 | 0.7454 |
| | | No | 24682.4 | Yes | 25913.2 | 82.071 | 670.7041 | 0.7405 |
| 527 × 2000 | IQL | Yes | 35217.2 | No | — | 10514.220 | 666.9160 | 0.7426 |
| | | No | 35188.0 | No | — | 10537.913 | 662.3462 | 0.7452 |
| | IAVRL | Yes | 23129.7 | Yes | 23978.1 | 75.858 | 670.5178 | 0.7405 |
| | | No | 32231.1 | Yes | 33850.8 | 109.154 | 673.9575 | 0.7370 |

Table 4. Performance improvement achieved by using prior knowledge compared with the case without using prior knowledge

| Gird | Algorithm | Performance improvement compared with the case without using prior knowledge (%) | | |
|---|---|---|---|---|
| | | Computation time reduce (%) | Return increase (%) | Execution time reduce (%) |
| 527 × 500 | IQL | 77.519 | 0.881 | 0.337 |
| | IAVRL | 37.851 | -2.011 | -1.628 |
| 527 × 1000 | IQL | 0.054 | 0.473 | 0.278 |
| | IAVRL | 32.326 | -1.078 | -1.057 |
| 527 × 1500 | IQL | 0.193 | 0.665 | 0.334 |
| | IAVRL | 27.070 | -0.643 | -0.662 |
| 527 × 2000 | IQL | 0.224 | 0.690 | 0.349 |
| | IAVRL | 30.504 | -0.510 | -0.475 |

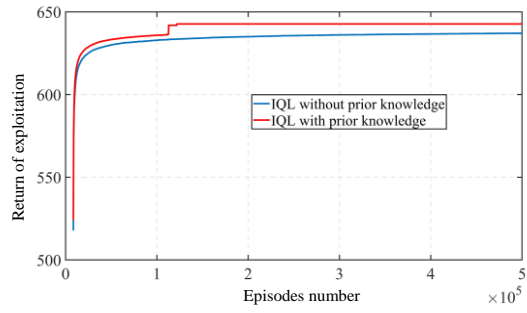

(a)

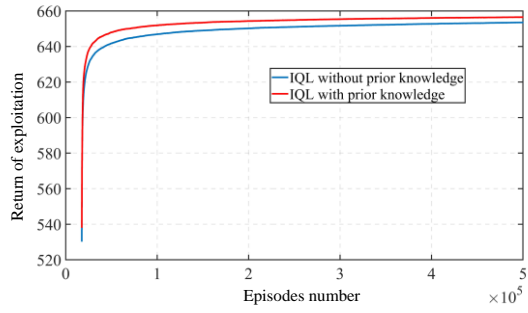

(b)

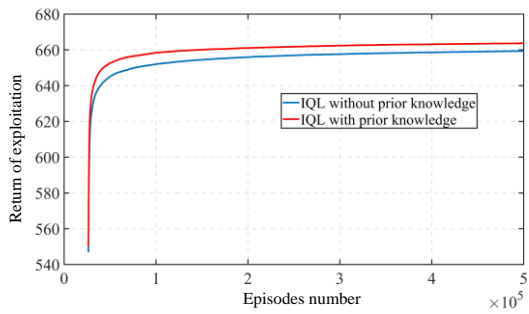

(c)

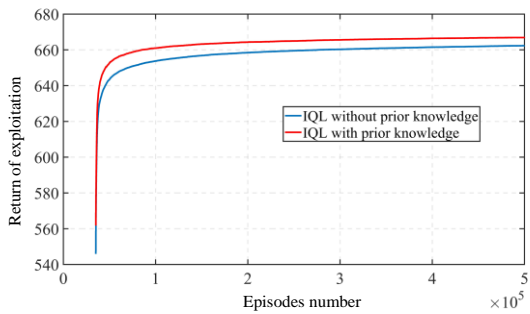

(d)

Figure 12. Return obtained by IQL with prior knowledge and IQL without prior knowledge through exploiting the learning experience after a successful exploration, where (a) is the case of a 527×500 grid, (b) is the case of a 527×1000 grid, (c) is the case of a 527×1500 grid and (d) is the case of a 527×2000 grid.

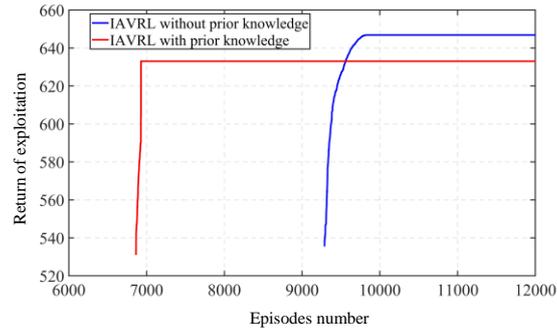

(a)

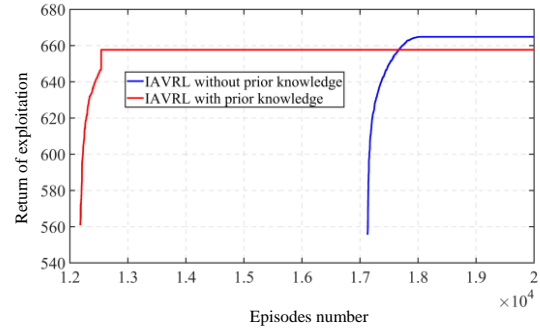

(b)

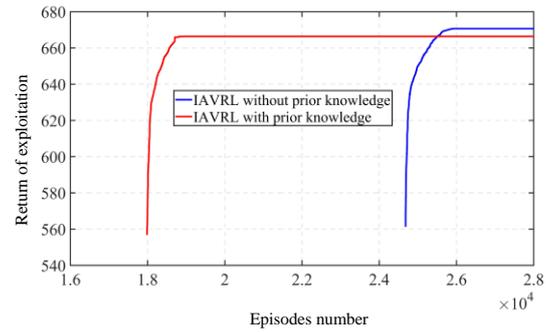

(c)

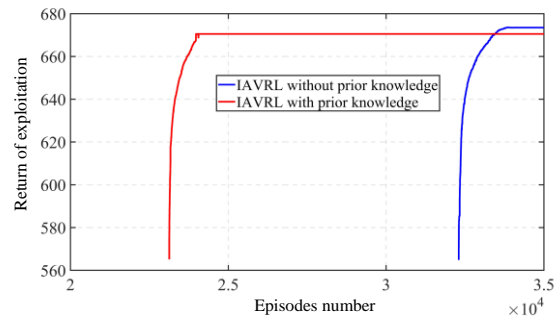

(d)

Figure 13. Return obtained by IAVRL with prior knowledge and IAVRL without prior knowledge through exploiting the learning experience after a successful exploration, where (a) is the case of a 527×500 grid, (b) is the case of a 527×1000 grid, (c) is the case of a 527×1500 grid and (d) is the case of a 527×2000 grid.

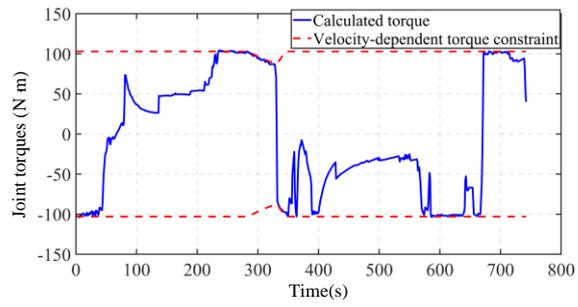

(a)

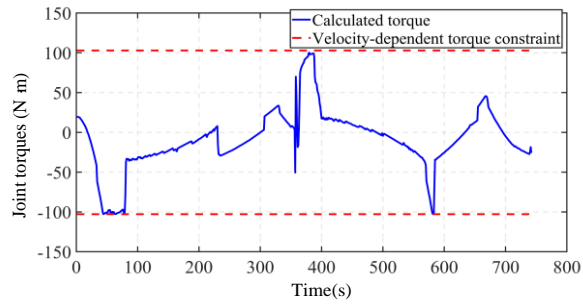

(b)

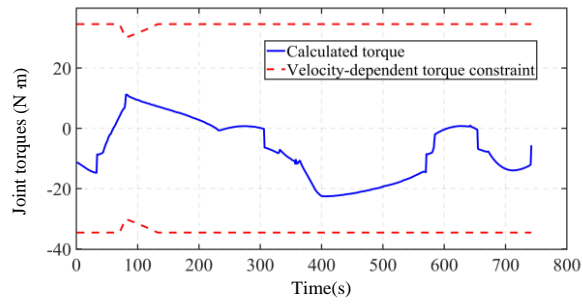

(c)

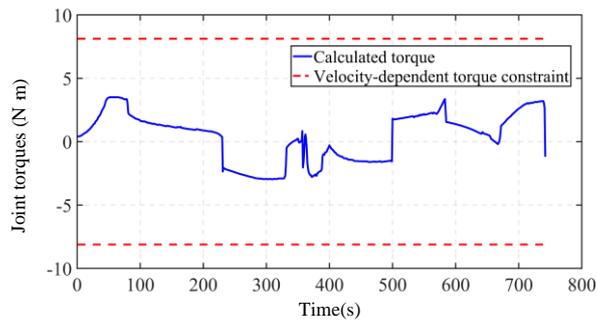

(d)

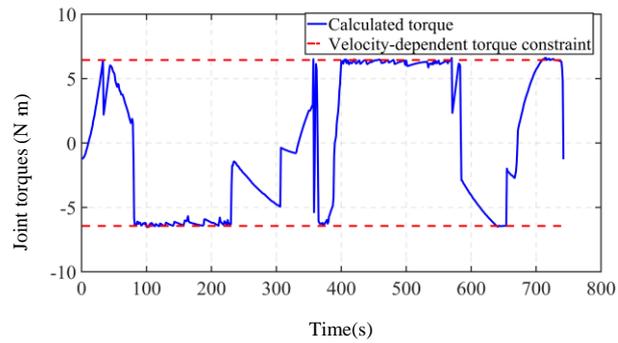

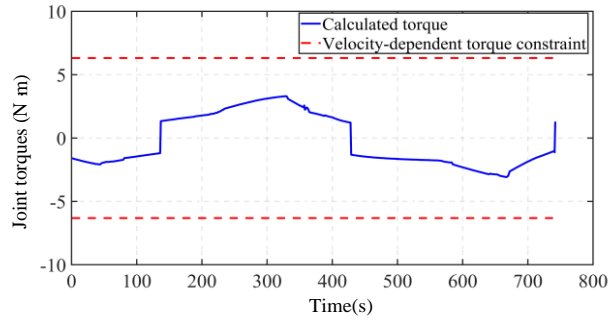

(f)

Figure 14. Calculated torques of the optimal trajectory obtained by IAVRL with prior knowledge in a 527×2000 grid case, where (a) is the torques of joint 1, (b) is the torques of joint 2, (c) is the torques of joint 3, (c) is the torques of joint 4, (d) is the torques of joint 5 and (e) is the torques of joint 6.

## 7. Conclusion

In this study, an improved Q-learning algorithm (IQL) and an improved action-value function reinforcement learning algorithm (IAVRL), have been proposed for the time-optimal path tracking problem. In order to construct the reinforcement learning states and decrease the learning dimension, a selective discrete method for discretizing the robotic task path is proposed. In order to improve the reinforcement learning convergence rate, an optimal trajectory obtained by direct planned method under conservative constraints is used as the prior knowledge to specify the initial Q value. Moreover, considering the limitation of the action-value function of Q-learning, a improved action-value function is proposed which is more suitable for solving the time-optimal path tracking problem. The effectiveness of the proposed algorithms are verified at a 6-DOF industrial robot. From a comparison experiment regarding reinforcement learning algorithm under conservative constraints, the feasibility of the proposed algorithms is verified, especially in the case of 527×2000 grid, the return of IQL can reach 96.9% of direct planned method NI-like and the return of IAVRL can reach 98.2% of NI-like. From a comparison experiment regarding reinforcement learning algorithm under velocity-dependent torque constraints, the effectiveness of the use of prior knowledge is verified, especially in the case of 527×500 grid, the computation time of IQL reduce 77.519% and the computation time of IAVRL reduce 37.851%. In addition, the effectiveness of the proposed method in solving the time-optimal path tracking problem under velocity-dependent torque constraints is verified, as the calculated constraints do not exceed the velocity-dependent torque constraints.

Despite the proposed algorithms provide feasible methods in solving the time-optimal path tracking problems, due to the limitation of grid, the proposed algorithms are just near optimal methods. In the future work, some other reinforcement learning algorithms can be considered, for example, the Deep Deterministic Policy Gradient (DDPG), which is an actor-critic, model-free algorithm for continuous action space and continuous state space. In addition, as the dynamic model which is used for planning is not a very precise model, the interaction with the real world instead of the dynamic model to find an optimal policy may also be considered in the future work.

## Acknowledge

This work is supported by National Science and Technology Major Project of China (No. 2015ZX04005006), Science and Technology Planning Project of Guangdong Province, China (2015B010918002), Science and Technology Major Project of Zhongshan city, China (No. 2016F2FC0006, 2018A10018).